\documentclass[11pt]{article}
\usepackage[preprint]{acl}
\usepackage{times}
\usepackage{latexsym}
\usepackage[T1]{fontenc}
\usepackage[utf8]{inputenc}
\usepackage{amsmath}
\usepackage{microtype}
\IfFileExists{inconsolata.sty}{\usepackage{inconsolata}}{}
\usepackage{graphicx}
\usepackage{booktabs}
\usepackage{tikz}
\usetikzlibrary{positioning, arrows.meta, fit, backgrounds, calc}
\usepackage{xcolor}
\usepackage{xurl}
\usepackage{url}
\usepackage[acronym]{glossaries}
\makenoidxglossaries % no makeindex run needed on arXiv
\newacronym{llm}{LLM}{Large Language Model}
\newacronym{tdg}{TDG}{Temporal Dependency Graph}
\newacronym{eat}{EAT}{Employment Appeal Tribunal}
\newacronym{acas}{ACAS}{Advisory, Conciliation and Arbitration Service}

\hypersetup{pdftitle={Time as Structure: A Temporal Backbone for Extraction and Reasoning over Legal Documents.},
  pdfauthor={Maryia Zhyrko, Suzan Verberne, Lifeng Han}}

\title{Time as Structure: A Temporal Backbone for Extraction and Reasoning
over Legal Documents}

\author{Maryia Zhyrko$^{1}$, Suzan Verberne$^{1}$, Lifeng Han$^{1,2}$ \\
  $^1$Leiden Institute of Advanced Computer Science (LIACS), Leiden University, NL \\
  $^2$Biomedical Data Sciences, Leiden University Medical Center, NL \\
  \texttt{mzhirko@gmail.com}}

\begin{document}
\maketitle

\begin{abstract}

Miss a filing deadline by one day and the claim is barred, however strong the case. Computing that deadline is rarely simple: the period runs from a triggering event, is counted by a statutory convention, and may be suspended by a mandatory conciliation window. We ask whether a language model should answer such questions directly, or read the document and leave the arithmetic to code. We extract dated facts and their dependencies into a temporal dependency graph and compute deadlines from it with a calendar-correct engine. On UK Employment Appeal Tribunal judgments the engine reproduces six of seven timeliness rulings, and matches the judges' own dates to the day. The strongest of four language models, asked the same cases, gets the arithmetic right and the answer wrong: in two of its seven responses its stated verdict contradicts its own working, and across the three GPT models six of twenty-one responses do, each time calling a late claim timely. To test the systems at scale we move the dismissal date across the statutory boundary, generating 427 cases whose answers are computed rather than annotated. On the cases both systems answer, the pipeline is right 90.2\% of the time against 61.2\% for direct answering. It answers two thirds of the cases, and counting its abstentions as errors it does not exceed direct answering (60.4\% against 64.6\%). The limit is extraction: on contracts the errors are almost never in the arithmetic, but in choosing which
event the period starts from.

\end{abstract}

\section{Introduction}
\label{sec:intro}

Miss a statutory filing deadline by one day and the claim is barred, whatever its merits. Computing that deadline is a chain: a period runs from a triggering event, is counted by a statutory convention, and may be paused for mandatory conciliation. Errors occur in practice: in the UK First-tier Tribunal (Immigration and Asylum Chamber), the share of disposals recorded as invalid or out of time rose from 3\% in 2023/24 to 9\% in the quarter to March 2026 \cite{mojtribunals}. Current tools handle the chain in one of two ways, and both leave a gap. Temporal taggers normalise each date on its own \cite{strotgen2010heideltime} and do not record which date is derived from which. Language models given the whole document must find the governing event, interpret the rule, do the arithmetic and report the result in one pass. They
order legal events reasonably well but struggle with nested legal language \cite{barale-etal-2025-lextime}, can state a conclusion their reasoning does not support \cite{wang2026knowlimitsfaithfulness}, and can be equally accurate while erring in opposite directions \cite{legalhallulens}. For deadlines the direction matters, because calling a late claim timely is the costly error. This paper introduces a method \footnote{The engine, graph representation, statutory rule packs and evaluation harnesses are released as an installable tool that prints the anchor, rule, arithmetic and confidence for every computation: \url{https://github.com/mzhirko/timebar}.} that splits the task: a model reads the document, and code does the computation. Our research question is whether \emph{separating extraction from deterministic computation gives more reliable and more verifiable deadline answers than direct answering by a language model, and where it fails}.

We compare two methods that receive the same inputs, a judgment and the governing statute section. In the \emph{structured} method, an extractor turns the judgment into a \gls{tdg}, a graph whose nodes are dated facts and whose edges record how one date depends on another. The statute passes through the same extractor and yields the period and the event it counts from. A deterministic engine binds the statute's anchor to a fact in the graph, applies calendar arithmetic and the \gls{acas} conciliation pause, and returns the deadline with the anchor, the rule and the intermediate dates. When it cannot tell which fact the anchor refers to, it abstains and names the cause. In the \emph{direct} method, a language model reads the same inputs and states the deadline, the verdict and its working in one response.

We compare the methods on the same items at three levels of strictness: the verdict (timely or late), the exact deadline date, and whether the working supports the stated verdict. The levels differ in practice. A claim presented 61 days late is still called late when the deadline is computed one day off, so a correct verdict can hide a wrong computation. Abstentions are reported separately, and we compare the methods both on the items both answer and over all items. Legal gold data is scarce, so we combine seven real tribunal rulings with two evaluations that need little annotation: a counterfactual sweep that moves the triggering date across the deadline and computes the new label from the statute, and a consistency check that needs no ground truth.

The structured method is exact on generated items, where direct accuracy falls as dependency chains lengthen. On UK Employment Appeal Tribunal judgments the engine reproduces six of seven timeliness rulings and every judge-stated date. On 427 perturbed cases it is right 90.2\% of the time on the cases both methods answer, against 61.2\% for direct answering, but it abstains on a third of the cases and is not more accurate over all of them. The strongest direct model computes correctly yet states a verdict that contradicts its own working in two of seven responses, and across the three GPT models in six of 21, each time calling a late claim timely. Our contributions are: 

\begin{itemize}
\item A structure-then-compute method for statutory deadlines: a temporal dependency graph, a deterministic engine with rules recovered from statute text, and abstention with a named cause, released as open source.

\item An evaluation design for settings with little gold data: working-level auditing, a counterfactual harness with computed labels, and a consistency measure that consumes no ground truth.

\item A comparison with five direct models that locates where each method fails: direct models in reporting their own computation, the structured method in choosing the event the period runs from.
\end{itemize}

The paper follows the order of the system. Section~\ref{sec:related} places the work in related research and Section~\ref{sec:data} describes the data. Section~\ref{sec:method} presents the method and the reason for each component. Section~\ref{sec:experiments} tests the components one layer at a time, first computation with given facts, then extraction, then both together on real and perturbed cases, and ends with a test that needs no ground truth. Section~\ref{sec:analysis} discusses what the results show and Section~\ref{sec:conclusion} answers the research question.

\section{Related Work}
\label{sec:related}

Temporal information extraction identifies and normalises dates, durations and intervals. HeidelTime \cite{strotgen2010heideltime} is a widely used rule-based system of this kind. It normalises expressions largely in isolation and does not record which event is derived from which. Temporal dependency parsing goes one step further: a narrative timeline can be represented as a dependency tree over events and time expressions \cite{kolomiyets-etal-2012-extracting}, and DocTime \cite{mathur-etal-2022-doctime}  parses document-level temporal dependency graphs end to end and releases an annotated graph dataset for contracts. In both, an edge states that one event precedes, follows or contains another. Our edges state something different: the offset the document gives and the counting convention it implies, so that a deterministic engine can compute one date from another and recompute it when the anchor moves. Their graphs answer which event came first. Ours answer how a date was derived, which is the link a deadline computation needs.

LexTime \cite{barale-etal-2025-lextime} shows that language models order legal events reasonably well and that nested legal language remains a bottleneck, and Sun et al. \cite{sun-etal-2026-sparse} show that ordering accuracy degrades sharply when explicit timestamps are replaced by relative expressions, which supports keeping anchor resolution and arithmetic outside the model. The task studied here extends beyond pairwise ordering to anchor identification, statutory rule application, calendar arithmetic and a deadline that can be compared against a court's own figure.

The work also relates to legal formalisation. Catala \cite{merigoux2021catala} is a programming language whose structure mirrors statutory text, and Monat et al. \cite{10.1007/978-3-031-57267-8_16} give a mechanised semantics for legal date arithmetic together with an analysis that detects counting ambiguities. Recent studies examine language model translation of law into Catala-like representations and document faithfulness failures in that route \cite{lorenzo2025taxlaw,wang2026knowlimitsfaithfulness}. Those studies evaluate generated code by similarity to a reference translation. We evaluate by execution, comparing the computed deadline against the one the court used.

SARA \cite{holzenberger-etal-2020-dataset} establishes statutory reasoning as a task, CLC-UKET \cite{xie-etal-2024-clc} predicts outcomes in the tribunal studied here, and LexNum \cite{zhang-etal-2025-legal} evaluates statute-guided numerical computation. Our perturbation design follows GSM-Symbolic \cite{gsm-symbolic} and replaces its templated answers with an oracle computed from the statute.

\section{Data and Task}
\label{sec:data}

The extraction experiments use contracts from the \texttt{en\_contracts} subset of Multi\_Legal\_Pile \cite{niklaus2024multilegalpile}. We sample 50 seeds, which return 45 documents, of which 41 are unique under exact hashing of the source text. The duplicates are one document appearing three times and two appearing twice. All extraction figures below are computed over the 41 unique documents.

The deadline task uses point-in-time versions of the Employment Rights Act 1996 s. 111, the Equality Act 2010 s. 123, and the corresponding \gls{acas} extension provisions, retrieved from legislation.gov.uk. Test cases are six \gls{eat} judgments decided in 2025 and 2026, retrieved from Find Case Law, the judgment service of The National Archives, and used under a transactional licence for computational analysis. We refer to them as Cases A--F. One judgment is evaluated under both statutes, giving seven statute-case rows. All six postdate the training cutoff of every model tested.

For the gold annotation, each row records the statutory anchor, the claim presentation date, the \gls{acas} certificate dates where an extension applies, the tribunal verdict, and the deadline or boundary date stated by the judge where the judgment states one. Two of the seven rows carry a judge-stated deadline and one carries a judge-stated boundary date. Every field is linked to a verbatim quotation and byte-checked against the source text by script.

To prevent leakage in the direct evaluation, we remove every sentence stating the deadline, the day count or the timeliness conclusion, and a leak check confirms that no deadline value survives. One leak cannot be removed: a party appeals what it lost, so the direction of the appeal can imply the first-instance verdict. We therefore treat deadline exactness and working-level consistency as more reliable than verdict accuracy alone, and we report where deference to the appeal direction affected a result.

\section{Method}
\label{sec:method}

This section describes the structured method and why it is built this way. Three properties of statutory deadlines shape the design.

First, a deadline is a derived date. It rarely appears in the judgment. It follows from another date by a rule, and that date may itself depend on others, such as a conciliation period. A list of normalised dates loses these dependencies. The \gls{tdg} keeps them as edges, so the anchor a deadline was computed from is explicit, and a corrected date propagates to every date derived from it (Section~\ref{sec:tdg}).

Second, the task mixes reading and counting. Deciding which event in the text is the \textit{effective date of termination} requires reading legal prose, which language models do well. Counting three calendar months less one day across a leap year requires exactness, which code guarantees. An extractor therefore builds the graph (Section~\ref{sec:extraction}) and a deterministic engine computes from it (Section~\ref{sec:engine}).

Third, a legal answer must be checkable, and a wrong deadline is worse than none. The engine returns its full working with every answer, and abstains with a named cause when it cannot tell which fact the statute's anchor refers to.

Figure~\ref{fig:architecture} shows how the components connect. Because extraction and computation are separate, every failure can be attributed to the component that produced it. Section~\ref{sec:eval} describes the direct baseline and how the two methods are scored.

\begin{figure*}[t]
\centering
\resizebox{\textwidth}{!}{%
\begin{tikzpicture}[
  font=\small,
  >={Stealth[length=2.2mm]},
  box/.style={draw, rounded corners=2pt, align=center, inner sep=5pt,
              minimum height=8mm, line width=0.5pt},
  input/.style={box, fill=black!4},
  extract/.style={box, fill=orange!12, draw=orange!65!black},
  graph/.style={box, fill=blue!10, draw=blue!55!black, line width=1pt},
  engine/.style={box, fill=blue!6, draw=blue!55!black},
  answer/.style={box, fill=green!9, draw=green!45!black, line width=1pt},
  abst/.style={box, fill=yellow!16, draw=yellow!55!black},
  direct/.style={box, fill=black!3, draw=black!45, dashed},
  lbl/.style={font=\scriptsize\itshape, text=black!60, align=center},
  flow/.style={->, line width=0.6pt},
  dflow/.style={->, line width=0.6pt, dashed, draw=black!50},
]

% ---- inputs
\node[input] (judg) {\textbf{Tribunal judgment}\\\scriptsize redacted or full text};
\node[input, below=8mm of judg] (stat) {\textbf{Statute section}\\\scriptsize point-in-time text};

% ---- extraction
\node[extract, right=16mm of judg, yshift=5mm] (llm)
  {\textbf{LLM extractor}\\\scriptsize single prompt, JSON facts + relations};
\node[extract, below=3mm of llm] (rule)
  {\textbf{Rule-based extractor}\\\scriptsize HeidelTime + spaCy + graph builder};

\coordinate (jsplit) at ($(judg.east)+(6mm,0)$);
\draw[line width=0.6pt] (judg.east) -- (jsplit);
\draw[flow] (jsplit) |- (llm.west);
\draw[flow] (jsplit) |- (rule.west);
\node[lbl, above=1mm of llm] {either extractor};

% ---- graphs
\node[graph, right=16mm of llm, yshift=-4mm] (casetdg)
  {\textbf{Case TDG}\\\scriptsize dated facts,\\\scriptsize typed edges};
\node[graph, below=20mm of casetdg] (ruletdg)
  {\textbf{Rule specification}\\\scriptsize period, anchor concept,\\\scriptsize counting convention};

\coordinate (esplit) at ($(llm.east)+(6mm,0)$);
\draw[line width=0.6pt] (llm.east) -- (esplit);
\draw[line width=0.6pt] (rule.east) -| (esplit);
\draw[flow] (esplit) -- ++(4mm,0) |- (casetdg.west);
\draw[flow] (stat.east) -- ++(8mm,0) |- (ruletdg.west);
\node[lbl, below=2mm of ruletdg]
  {statute passes through the same extractor,\\convention confirmed once by hand};

% ---- engine internals
\node[engine, right=20mm of casetdg, yshift=6mm] (match)
  {\textbf{1. Anchor matcher}\\\scriptsize bind statutory concept to a case fact};
\node[engine, below=3mm of match] (calc)
  {\textbf{2. Calendar arithmetic}\\\scriptsize anchor $+$ period $-$ 1 day,\\\scriptsize real months and leap years};
\node[engine, below=3mm of calc] (pause)
  {\textbf{3. Conciliation pause}\\\scriptsize suspend Day A to Day B,\\\scriptsize one-month floor};

\draw[flow] (casetdg.east) -- ++(6mm,0) |- (match.west);
\draw[flow] (ruletdg.east) -- ++(6mm,0) |- (match.west);
\draw[flow] (match) -- (calc);
\draw[flow] (calc)  -- (pause);

\begin{scope}[on background layer]
  \node[draw=blue!55!black, rounded corners=3pt, fit=(match)(calc)(pause),
        inner sep=4mm, fill=blue!3,
        label={[lbl, yshift=-1mm]north:deterministic engine}] (enginebox) {};
\end{scope}

% ---- outputs
\node[answer, right=16mm of match] (answer)
  {\textbf{Deadline and verdict}\\\scriptsize with anchor, rule, arithmetic\\\scriptsize and confidence};
\node[abst, below=16mm of answer] (abstain)
  {\textbf{Abstention}\\\scriptsize named cause,\\\scriptsize confidence 0.00};

\draw[flow] (enginebox.east) |- (answer.west)
  node[lbl, pos=0.75, above] {anchor binds};
\draw[flow] (enginebox.east) |- (abstain.west)
  node[lbl, pos=0.72, below, align=center] {no candidate, or\\candidates conflict};

% ---- direct baseline path
\node[direct, above=14mm of casetdg] (dbase)
  {\textbf{Direct baseline}\\\scriptsize model reads both inputs, answers in one pass};
\node[direct, right=16mm of dbase] (dout)
  {\textbf{Deadline and verdict}\\\scriptsize no abstention available};

\coordinate (dup) at ($(judg.north)+(0,10mm)$);
\draw[line width=0.6pt, dashed, draw=black!50] (judg.north) -- (dup);
\draw[dflow] (dup) -| (dbase.west);
\draw[dflow] (dbase.east) -- (dout.west);
\node[lbl, above=1mm of dbase] {comparison condition};

\end{tikzpicture}%
} \caption{System architecture. Either extractor produces a case TDG, and the statute section passes through the same pipeline to yield the rule specification. The deterministic engine binds the statutory anchor concept to a case fact, applies calendar arithmetic and the conciliation pause, and returns either a deadline with its full working or an abstention with a named cause.
The dashed path is the direct baseline, which receives the same two inputs and answers in one pass.}
\label{fig:architecture}
\end{figure*}

\subsection{Temporal dependency graph}
\label{sec:tdg}

Nodes in a \gls{tdg} store an entity, a semantic role in \{\texttt{START}, \texttt{END}, \texttt{DURATION}, \texttt{CONTAINS}\}, and a date normalised to ISO~8601. Edges are typed. An \texttt{additive} edge states that the target date is computed from the source date by a stated period. An \texttt{ordering} edge states sequence with no offset. The type distinction determines what the engine can execute: only additive edges with a resolved source date support arithmetic.

An additive edge carries the period twice. \texttt{constraint\_expr} holds it in the units the source text uses, as an expression such as \texttt{"e2 = e1 + 3 months"} with the duration node valued \texttt{P3M}, and \texttt{delta\_days} holds an integer approximation used only for sorting and for cheap consistency checks. \texttt{constraint\_expr} is the authority. Months and years are never converted to a day count at extraction time, and the engine applies them as calendar months and years through \texttt{relativedelta}, keeping years, months and days apart rather than collapsing them. So when an anchor date changes, a three-month edge still ends three calendar months later less one day under the \textit{beginning with} convention, at whatever month length the new anchor implies, and the stale \texttt{delta\_days} value plays no part in the computation.

\subsection{Extraction}
\label{sec:extraction}

The rule-based pipeline runs four steps. HeidelTime extracts and normalises temporal expressions. A role classifier walks the spaCy \cite{honnibal2020spacy} dependency tree from the date token to the head verb of its clause and matches lemmas, so that \textit{began}, \textit{beginning} and \textit{starts} all map to \texttt{START}, with a special case for legal genitives such as \textit{Decision of 18 March 1992}. An entity linker attaches dates to entities using named entity recognition and coreference resolution. The graph builder collects \texttt{START}--\texttt{END} pairs sharing an entity, compares the calendar gap against every stated duration in the document, and types the edge additive when a duration matches and ordering otherwise.

HeidelTime resolves incomplete expressions such as \textit{by 31 March} against a document creation time, which this corpus does not supply. A fixed global value places such dates outside the document's own period, and in one 1984 instrument produced a date in 2020. We therefore derive a per-document creation time from the document's earliest explicit date, ignoring bare years because citation years are not document dates. This resolves 43 of the 45 documents.

The \gls{llm} pipeline replaces all four steps with a single prompt. The model receives the raw document and returns JSON containing facts and relations. The prompt supplies trigger phrases for additive dependencies (\textit{within X days of}, \textit{no later than X days from}), a worked mapping from \textit{within 30 days of the effective date} to an additive edge with \texttt{delta\_days: 30}, and counterexamples preventing monetary amounts and percentages from being typed as durations. Post-processing removes self-loops, downgrades additive edges whose source sentences contain no trigger phrase, and downgrades edges whose target already carries a fixed calendar date.

\subsection{Deterministic deadline engine}
\label{sec:engine}

The engine takes a statute graph and a case graph. A matcher binds the statutory anchor concept to a fact in the case graph. The engine then computes the deadline using real month lengths and leap years.

For periods worded \textit{beginning with} an anchor, the UK convention makes a three-month period end three calendar months after the anchor, less one day. The engine implements this as anchor $+$ 3 months $-$ 1 day, with month-end rounding handled explicitly. Sections 207B ERA and 140B EqA suspend the clock between Day~A, the date of contact with \gls{acas}, and Day~B, the date the certificate is issued, and guarantee at least one month after Day~B. The engine implements the suspension, the one-month floor, the re-anchoring rule, and the precondition that Day~A falls within the primary period. All conventions are covered by unit tests, and Figure~\ref{fig:mechanics} in Appendix~\ref{app:mechanics} shows one computation with every quantity the engine reports.

The rules are not hard-coded per case. The statute section is passed through the same extraction pipeline as any other document, which returns the period and the anchor concept it counts from. Only the counting convention is confirmed by hand once per statute and recorded in a declarative specification.

When the matcher finds no candidate, or several conflicting candidates, the engine returns \texttt{INDETERMINATE} with a machine-readable cause. We count these abstentions separately from wrong answers throughout.

\subsection{Baselines and evaluation}
\label{sec:eval}

The direct baseline supplies each model with the redacted judgment, the governing statute section and the conciliation rule, and asks for a deadline, a verdict and the arithmetic. This is more context than a retrieval system would supply, so the baseline upper-bounds retrieval-based approaches on this task. Models play two roles, and each model is not a separate pipeline. As direct baselines we run \texttt{gemma4:e4b} and \texttt{llama3.1:8b} locally and \texttt{gpt-5.4-nano}, \texttt{gpt-5.4-mini} and \texttt{gpt-5.4} through the API. Inside the structured method a model only extracts the graph, and the same engine computes every answer: \texttt{gemma4:e4b} and \texttt{gpt-5.4-mini} extract on the tribunal cases, and \texttt{gemma4:e4b} and \texttt{llama3.1:8b} on the perturbed cases. Each table names the models it uses. Prompts are frozen across models, temperature is 0, and all raw responses are archived. Reported numbers are produced by script and not counted by hand.

Controlled items are scored by calendar value and not by surface form. Real cases are scored by verdict, by exact deadline where the judgment states one, and by agreement between the working shown and the fields emitted. One method is better than another if it is right more often on the same items. Because the structured method can abstain, we report its coverage and compare accuracy both on the items both methods answer and over all items, counting abstentions as
errors.

\section{Experiments and Results}
\label{sec:experiments}

The experiments test the method one layer at a time and then end to end, so that each failure can be traced to a layer. Each subsection answers one question. (1)~With the facts given, is the computation right, and how do direct models degrade as dependency chains lengthen? (2)~Can extractors recover the dependencies the engine needs from real legal prose? (3)~On real tribunal cases with verified facts, does the engine reproduce the court's ruling, and how do direct models do on the same cases? (4)~How does the full pipeline perform with no human input? (5)~At larger scale, does each method follow the deadline when the triggering date moves? (6)~Without ground truth, are a model's answers at least consistent with one another?

\subsection{Computation with given facts}

The controlled benchmark separates reasoning from extraction. It contains 210 generated items in seven categories of 30, with anchors drawn from a stress list of 31 January, 29 and 30 January, 29 February 2024, and month ends including 31 August and 30 November. Deadline items give an anchor, a period and a margin from $\{-7,-4,-3,-1,1,3,4,7\}$ days. Cascade items give a root date and a chain of two or three offsets, then replace the root, so the system must propagate the correction. Ground truth is computed from the calendar at generation time.

\begin{table}[t]
\centering
\small
\begin{tabular}{lrrrr}
\toprule
Category & Struct. & Naive & gemma & llama \\
\midrule
Digits period & 100 & 50 & 67 & 63 \\
Words period & 100 & 70 & 57 & 40 \\
Vague period & 100 & 67 & 47 & 23 \\
2 links, one doc. & 100 & 37 & 93 & 0 \\
2 links, cross-doc. & 100 & 40 & 97 & 10 \\
3 links, one doc. & 100 & 20 & 97 & 3 \\
3 links, cross-doc. & 100 & 7 & 90 & 0 \\
\midrule
Overall items & 100 & 41 & 78 & 20 \\
\bottomrule
\end{tabular}
\caption{Percentage of items fully correct in the controlled benchmark. A cascade item is correct only when every linked date is correct.}
\label{tab:controlled}
\end{table}

The structured method is exact in every category, and direct accuracy falls as the period becomes less explicit and the chain lengthens (Table~\ref{tab:controlled}). Gemma propagates corrections in most cases, with per-link accuracy of 97\%, 96\% and 93\% across the three cascade steps. Llama restates the original dates instead of applying the correction and falls from 11\% at the first link to 3\% at the third. The calendar-naive baseline reaches 41\% overall and still exceeds llama's 20\%, which indicates that applying crude arithmetic to the correct anchor outperforms fluent output that does not propagate.

\subsection{Extraction from contracts}

We ran both extractors over the same 41 contracts and audited every edge the engine could execute. A computable edge is additive, has a resolved source date, and carries an offset. Ordering edges claim no offset by definition, and interval edges in this corpus connect nodes with no date.

\begin{table}[t]
\centering
\small
\begin{tabular}{lrr}
\toprule
 & Rule-based & LLM \\
\midrule
Facts extracted & 592 & 288 \\
\quad resolving a date & 486 & 95 \\
\quad usable nodes & 188 & 50 \\
\midrule
Dependency edges & 11 & 92 \\
\quad additive & 9 & 33 \\
\quad ordering & 2 & 40 \\
\quad interval & 0 & 19 \\
Documents with an edge & 7/41 & 30/41 \\
\midrule
Computable edges & 9 & 10 \\
\quad offset matches duration & 9/9 & 12/13 \\
\quad surviving audit & 1 & 3 \\
\bottomrule
\end{tabular}
\caption{Extraction over 41 unique contracts. A usable node has a resolved calendar date, is not a bare citation year, and carries a nameable entity. Every computable edge on both sides was audited against its source sentences. We report precision on produced edges. Recall would require dependency-level annotation the corpus does not carry.}
\label{tab:extract}
\end{table}

The rule-based pipeline resolves 486 dates against the \gls{llm}'s 95 and produces 11 edges against 92, with edges on 7 documents against 30 (Table~\ref{tab:extract}). Normalising individual expressions accurately therefore does not by itself recover dependency structure. The \gls{llm}'s advantage also narrows once the set is restricted to edges the engine can
execute. Of its 92 edges, 31 have a resolved date at neither endpoint, so a relation is recovered but no arithmetic can be performed along it.

The audit identifies what the surviving edges have in common. One of the nine rule-based edges survives, a single-sentence clause reading \textit{extended for a period of three months from 15 March 1986}. Three of the ten \gls{llm} edges survive, and all three contain an explicit trigger phrase naming the anchor concept in the same sentence, for example \textit{within one month after the entry into force of this Agreement}. A fourth carries a real offset attached to the wrong event, counting twenty-one days from the date of a letter instead of from the entry into force the clause names.

Offsets are internally consistent in almost every case: 9 of 9 rule-based offsets match their normalised duration, and 12 of 13 for the \gls{llm}. The single exception disagrees with itself twice, reading \textit{three months} in its expression, carrying a ninety-day duration on its endpoint, and recording thirty in its offset field. The remaining failures are anchor selection and entity identity. Over half the rule-based entities are unusable, most of them clause fragments, and the graph builder pairs a start with an end on a shared entity, so a noun phrase lifted out of context leaves nothing to pair. Both extractors make the same error on one document, anchoring \textit{three years after the entry into force} to a different date stated nearby, which suggests anchor selection is a property of the task and not of either implementation.

\subsection{Engine and direct models on tribunal cases}
\label{sec:baseline}

We first test the engine on human-verified facts, which isolates it from extraction. It reproduces six of seven tribunal verdicts, both judge-stated deadlines, and the judge-stated boundary date in the continuing-act case. The seventh row states an in-time conclusion without reciting the conciliation dates that produced it. We call such rows document-bound. The engine computes the primary deadline for that row exactly and reports the missing input.

\begin{table}[t]
\centering
\small
\begin{tabular}{lrrr}
\toprule
Model & Verdicts & Deadlines & Mean $|\Delta d|$ \\
\midrule
gemma4:e4b & 1/7 & 0/6 & 87.2 \\
gpt-5.4-nano & 3/7 & 0/6 & 77.3 \\
gpt-5.4-mini & 1/7 & 0/6 & 99.0 \\
gpt-5.4 & 3/7 & 3/6 & 50.5 \\
\midrule
Engine, gold facts & 6/7 & 2/2$^{\dagger}$ & --- \\
\bottomrule
\end{tabular}
\caption{Direct \gls{llm} baseline on redacted judgments, with mean absolute deadline error in days. The model deadline denominator is six, not two: models are scored against the reference deadline on every row for which one is computable, which is the six rows with complete inputs. The seventh (document-bound) row recites neither its deadline nor the conciliation dates that produced it, so no reference deadline exists for it and it is excluded. $^{\dagger}$The engine is held to the stricter standard of the judgments' own figures, so it is scored only on the two judge-stated deadlines. It additionally reproduces the one judge-stated boundary date.}
\label{tab:llm}
\end{table}

Deadline exactness improves with model scale and verdict accuracy does not (Table~\ref{tab:llm}). The strongest model applies the minus-one-day convention correctly in all seven responses and produces three of six computable deadlines exactly. In two rows its emitted verdict field contradicts its own correct working, and across the GPT family six of 21 responses contain this divergence, all in the direction where the working concludes out of time and the field says in time. The local model shows none. The smaller models fail the convention itself, applying it in 0 of 7, 0 of 7 and 1 of 7 responses, and none below the frontier applies the s. 207B pause correctly anywhere.

One alternative reading of the divergence is available. The verdict field precedes the arithmetic in the emitted JSON, so this data does not distinguish a model that commits early and rationalises afterwards from one that computes and then misreports. The conclusion holds under both readings: the emitted field is not usable without the working.

\subsection{Full pipeline without human input}

We ran the full chain with no human in the loop, using two extractors over redacted and full-text judgments with the same engine, thresholds and fallback in every condition.

Extraction recall is high. The strongest condition finds the gold anchor in six of seven rows, all seven presentation dates, all four available \gls{acas} Day A dates and all three Day B dates. The engine nevertheless abstains often, because the matcher does not align a statutory concept such as \textit{effective date of termination} with an extracted label such as \textit{dismissal}. The bottleneck therefore moves from extraction to semantic binding.

The matcher uses five statute-grounded mechanisms to make this binding: anchor aliases derived from the statutes' own vocabulary, a stoplist of procedural events, claim-presentation cues with an earliest-date tie-break, a conflict gate that surfaces disagreeing anchor candidates, and an alias tier so that weak lexical similarity cannot outrank statutory vocabulary. An ablation over the same 28 graphs, with no re-extraction and no new model calls, shows that these mechanisms raise correct bindings from 6 of 10 to 12 of 14 across the affected tables.

Across the 28 cells the engine answers 14 and abstains on 14. Every abstention carries confidence 0.00 and a named cause, and no deadline is fabricated in any cell. Twelve of the 14 answers are correct verdicts, and both errors are the same document-bound row.

Verdict agreement alone overstates this result, so we score the same 14 answers at the three levels of strictness introduced in Section~\ref{sec:intro} (Table~\ref{tab:tiers}). Twelve have the correct verdict. Nine also have the correct day count, and the same nine pass the strictest level, which requires the correct anchor, rule, action date and computation: on this set, every answer with the right day count also used the right anchor. The other three are correct only because the margin is large. Two are the same case under two conditions, where the anchor is one day early and the verdict does not change. In the third, the extractor took a grievance for the claim and computed the deadline from the wrong anchor, yet the verdict is still right. The 28 direct baseline responses contain no abstentions, and every failure is stated with confidence.

\begin{table}[t]
\centering
\small
\begin{tabular}{lr}
\toprule
Criterion & Answers \\
\midrule
Correct verdict & 12/14 \\
\quad + correct day count & 9/14 \\
\quad + correct anchor, rule and action date & 9/14 \\
\bottomrule
\end{tabular}
\caption{The 14 answered cells at three levels of strictness. The criteria are nested, so each line is a subset of the one above. The gap between the first and second lines is the three answers that are correct by margin.}
\label{tab:tiers}
\end{table}

\subsection{Boundary tracking on perturbed cases}

The gold set supports a taxonomy of failures and not error rates. To obtain computed ground truth at scale we perturb the anchor date of each real judgment and recompute the label through the engine.

For each gold case we sweep the anchor across $k \in [b-30, b+30]$ days, where $b$ is the engine-computed boundary offset for that case, so every sweep straddles its own statutory boundary. This gives $7 \times 61 = 427$ items balanced 31/30 timely to late within each case. A self-check verifies that every sweep is monotone with exactly one verdict flip at the boundary the
statute prescribes.

The anchor is shifted together with every earlier date in its narrative, so that each perturbed judgment stays internally consistent while the litigation timeline stays fixed, and a generation check rejects any text in which a reference to the original date survives (Appendix~\ref{app:counterfactual}).

The engine generates the labels and is not scored against them. Because the same engine also computes the pipeline's answers, this experiment cannot detect an error in the rule implementation: a misread statute would be reproduced in both. It tests whether extraction and anchor binding recover inputs consistent with the implemented rule, and it measures text-reading systems against statute-computed truth. The legal correctness of the rule itself rests on the
seven gold rows, where the engine reproduces the tribunal's own dates.

\begin{table}[t]
\centering
\small
\setlength{\tabcolsep}{4pt}
\begin{tabular}{lrrrrr}
\toprule
Condition & Cov. & Acc$_{\text{all}}$ & Acc$_{\text{ans}}$ & Flips & 1-flip \\
\midrule
gemma direct & 1.00 & 0.646 & 0.646 & 68 & 1/7 \\
llama direct & 1.00 & 0.602 & 0.602 & 117 & 0/7 \\
gemma pipeline & 0.67 & 0.604 & \textbf{0.902} & 24 & 3/7 \\
llama pipeline & 0.41 & 0.335 & 0.822 & 18 & 0/7 \\
\bottomrule
\end{tabular}
\caption{427 perturbed items, majority class 0.508. A system tracking the statute flips its verdict exactly once per sweep. Acc$_{\text{all}}$ counts abstentions as errors and Acc$_{\text{ans}}$ is over answered items. Coverage differs between conditions, so the matched comparison in the text is the fair one.}
\label{tab:cf}
\end{table}

\begin{table}[t]
\centering
\small
\setlength{\tabcolsep}{4pt}
\begin{tabular}{lrrr}
\toprule
Case & $n$ & pipeline & direct \\
\midrule
Case A (s.111) & 60 & 0.983 & 0.467 \\
Case A (s.123) & 60 & 0.983 & 0.567 \\
Case B        & 49 & 1.000 & 0.510 \\
Case C        & 55 & 1.000 & 0.982 \\
Case D        & 41 & 0.415 & 0.390 \\
Case E        & 14 & 1.000 & 0.857 \\
Case F        &  7 & 0.714 & 0.857 \\
\midrule
Mean over cases  & & 0.871 & 0.661 \\
SD over cases    & & 0.227 & 0.232 \\
Mean over items  & & 0.902 & 0.612 \\
\bottomrule
\end{tabular}
\caption{Matched per-case comparison, gemma extractor: the items answered by both the pipeline and the direct model ($n=286$ of 427). Cases are anonymised as A--F; Case A is evaluated under both statutes. The case is the unit of analysis, so the case-level mean weights every case equally and differs from the item-level mean of Table~\ref{tab:cf}, which is dominated by the cases the pipeline answers densely.}
\label{tab:cf-percase}
\end{table}

Both direct models answer every item and sit above the majority class of 0.508 by 14 and 9 points (Table~\ref{tab:cf}). The pipeline answers fewer items and reaches 90.2\% on those it answers, and 98.3\% on the subset where the extracted anchor equals the true shifted anchor ($n=234$). On the items both gemma conditions answered, the pipeline reaches 90.2\% against 61.2\% direct ($n=286$, disagreements 93 to 10); for llama, 82.2\% against 54.6\% ($n=174$, 62 to 14). We report no significance test on these item counts. The 61 items within a case are near-identical by construction, so a system that binds the wrong anchor on the base case usually fails on all 61: the statistical unit is seven cases, not 427 items, and an item-level test would overstate confidence. Table~\ref{tab:cf-percase} therefore gives the comparison per case.

Abstention does not track item difficulty. Direct accuracy is \emph{higher} on the items the pipeline abstained from (71.6\%, $101/141$, for gemma; 64.0\%, $162/253$, for llama) than on the items it answered (61.2\% and 54.6\%). This follows from what produces an abstention. Every abstention is raised before the arithmetic, by the matcher or by a missing fact, and never by the computation: for gemma, 77 of 141 are \textit{no candidate} for the statutory anchor concept and 64 are the conflict gate declining to choose between distinct-dated candidates. For llama the same two causes account for 174 and 69 of 253, with a further 10 where an anchor bound but no action fact followed it. Coverage therefore follows anchor binding, which is independent of whether a text-reading model finds the item easy. On this set the two run opposite ways: the pipeline answers almost every item of the two Case A rows and of Case B, where direct scores 0.467, 0.567 and 0.510 (Table~\ref{tab:cf-percase}), and abstains on most items of Cases E, F and D, where direct scores 0.70 ($33/47$), 0.69 ($37/54$) and 0.80 ($16/20$) on
the abstained items.

Over all 427 items the pipeline does not exceed direct answering: gemma direct reaches 0.646 against the pipeline's 0.604, because coverage is bounded by extraction recall. The result is the asymmetry. Where the anchor binds the pipeline is near-exact, where it does not the pipeline abstains with a named cause, and on rows where the extracted anchor is wrong the pipeline sits at chance (0.538). Flip counts show the same pattern structurally: the direct models flip 68 and 117 times across seven sweeps, against 24 and 18 for the pipeline and 7 for a system tracking the statute.

\subsection{Consistency without ground truth}

Every measure above consumes an answer known in advance, which is what limits the evaluation to seven cases. We therefore add a measure that consumes none.

Six questions are put to a system about one case: the presentation date $p$, the effective deadline $d$, the last date $\ell$ on which the claim could still have been in time, the verdict $v$, the number of days late $\delta$, and whether presentation complied. Four constraints hold between the answers by arithmetic or definition, independently of the case: $\ell = d$ (K1), $\delta = p - d$ (K2), $(v = \textit{in\_time}) \iff (\delta \le 0)$ (K3), and $\textit{complied} \iff (v = \textit{in\_time})$ (K5). A system violating any of these has contradicted itself whether or not we know which answer is correct. A fifth relation is entailed by K1 to K3 and is reported only as a redundancy check. K5 asks the same question as K3 with a different answer type, so a violation confined to K5 indicates sensitivity to phrasing. Each question is asked in its own call with no shared context, and because that choice is open to argument we also run a single-prompt condition identical in every other respect. Appendix~\ref{app:coherence} gives the constraint-level results.

\begin{table}[t]
\centering
\small
\setlength{\tabcolsep}{4pt}
\begin{tabular}{lrrr}
\toprule
Condition & Coverage & Incoherent & Verdicts \\
\midrule
llama3.1:8b, separate & 4/7 & 4/4 & 1/7 \\
llama3.1:8b, single & 6/7 & 6/6 & 4/6 \\
gemma4:e4b, separate & 5/7 & 4/5 & 4/6 \\
gemma4:e4b, single & 7/7 & 0/7 & 2/7 \\
gpt-5.4, separate & 7/7 & 4/7 & 5/7 \\
gpt-5.4, single & 7/7 & 0/7 & 4/7 \\
\bottomrule
\end{tabular}
\caption{Consistency on the seven gold rows. Coverage counts cases where all six answers parsed and incoherence is scored over those. Verdicts are scored against the tribunal's ruling over the rows whose verdict field parsed, which is a larger set, so the two columns are not nested.}
\label{tab:coherence}
\end{table}

Asked the same case six times, gpt-5.4 contradicts itself on four of seven at full coverage (Table~\ref{tab:coherence}). Answering in one request removes self-contradiction for both capable models, and verdict accuracy against the tribunal's ruling falls in both, from five of seven to four of seven for gpt-5.4 and from four of six to two of seven for gemma. The engine is not scored here. Its six fields are projections of one computation, so it cannot violate a constraint, and we treat that as a property of the architecture.

The violations locate the failure. For gpt-5.4, K3 and K5 are 0 of 7 and all four violations are K1 and K2, so the verdict never disagrees with its own day count while the dates and the subtraction do. This differs from the divergence in Section~\ref{sec:baseline}, where the reported answer was unfaithful to correct working. Here the working is unstable across askings and the reporting is exact. For llama in the single-prompt condition, K2 is 6 of 6 and K5 is 0 of 6: with both operands present in its own output it performed the subtraction wrongly every time while never contradicting its own conclusion.

\section{Analysis}
\label{sec:analysis}

Verdict accuracy and deadline exactness each miss a failure the other detects: verdict-level scoring passes a wrong computation when the margin is large (three of the twelve correct pipeline verdicts), and value-level scoring passes a right number produced by two errors that cancel (once). Only comparing the working against the emitted fields detects the divergence class, which is present in the strongest model and absent in the weakest.

The structured system differs from direct answering in several respects at once: an explicit graph, deterministic execution, statute-grounded anchor matching, a conflict gate and abstention. The results therefore support separating extraction from execution, and they do not isolate the contribution of the graph itself. On the tribunal task the engine consumes a small set of dated facts and takes the period from the statute graph, so a flat structured record with the same engine may perform comparably there. The graph does distinct work in propagation, where a corrected root must reach every dependent date (the cascade items). A flat-record baseline on both settings is the direct test of the representation, and we leave it to future work.

Self-consistency is also not a sign of correctness. Removing self-contradiction did not improve accuracy in either model where it took effect, and verdict accuracy fell in both, so self-consistency is not usable as a confidence signal. What the deterministic layer provides is not higher accuracy but a located cause: when the verdict is the last line of a computation, a wrong verdict is attributable to an anchor, a rule or an input.

The main open problem is anchor selection. Continuing acts, rejected factual contentions and adjacent procedural events produce competing plausible anchors. The direct models select the wrong one, and the extractors label the wrong one, in one condition taking the claimant's rejected contention as the dismissal date, where the matcher's confidence of 0.24 flagged the binding as weak. In the continuing-act case, the appellate court described the tribunal's own scoping as clearly open to it, which makes the contrary view arguable. Better extraction does not resolve a case where the anchor depends on a legal conclusion. The conflict gate presents the candidates and their computations instead of selecting silently.

\section{Conclusion}
\label{sec:conclusion}

We introduced a structure-then-compute method for statutory deadlines and asked whether separating extraction from deterministic computation gives more reliable and more verifiable answers than direct answering, and where it fails. On the cases it answers, it does: it is exact on generated items, reproduces six of seven tribunal rulings and every judge-stated date, and on 427 perturbed cases is right 90.2\% of the time against 61.2\% for direct answering, with every answer carrying its anchor, rule and arithmetic. Direct answering fails in reporting: the strongest model computes correctly and then states the opposite verdict, always in the costly direction. The structured method fails in choosing the event the period runs from, so it abstains on a third of the cases and is not more accurate over all of them. The open problem is therefore binding statutory anchor concepts to the right facts, and where that choice is a legal judgment, presenting the candidates and their consequences instead of choosing silently.

\section*{Limitations}

The legal gold set has seven rows, which supports a taxonomy of failure mechanisms and not error rates, so we report counts instead of percentages. The counterfactual sweep raises the item count to 427 with computed ground truth, but its statistical unit remains seven cases. The statutes come from one jurisdiction and one procedural domain. Establishing how far the method generalises requires more data: further statutes, courts, jurisdictions and deadline types, with annotated extraction recall. We are building such evaluation sets and will report them in upcoming publications. The controlled benchmark is synthetic, with prose explicit enough that extraction is trivial, so it isolates the reasoning component by design. Published time-limit judgments are also not a random sample of time-limit disputes, because a case turning on the point is more likely to be reported when the claim was late, so the in-time direction is
represented by one real case and by the sweep.

The engine, its counting conventions and the rule specifications were fixed before the tribunal evaluation. Validation of the matcher on held-out cases beyond the seven studied here is still required, as is a flat-record ablation that separates the graph representation from deterministic execution. The method helps only where there is arithmetic to verify: on pairwise event ordering \cite{barale-etal-2025-lextime} it loses to direct answering (Appendix~\ref{app:negative}).

In a separate pilot, a legal professional reviewed the released tool on three Dutch and EU immigration documents. All 18 extracted rows were correct, but six legally relevant dates were missing, and the tool's recall check flagged only two of them. The silent misses were month-and-year dates, a second event on a date already listed, and a date written with a Dutch month name. The review also found a deadline counted from a decision date where the statute counts from
notification, without the assumption being stated. This is one reviewer on three documents, so it shows types of failure, not rates.

On extraction we report precision on produced edges and no recall, because the contracts carry no dependency-level annotation. Audit verdicts are judgements by a single assessor, applying a scoring rule fixed before the audit, and are under independent review. \gls{llm} extraction varies between runs at temperature 0, so all counts come from a single named run. Only explicit temporal expressions are detected.

Coverage in the consistency measure is not missing at random. All parse failures fall on three cases, two of which carry the anchors that are hardest elsewhere in our results, so accuracy over answered items is computed on an easier subset. We exclude llama from the single-prompt comparison, because its two conditions differ by a lost case and a changed generation task as well as by shared context.

\section*{Ethical Considerations}

The system computes statutory deadlines and does not give legal advice. Its
outputs are intended to be checked against the source documents it cites, and
abstention is deliberate: where selecting the anchor is a legal judgement rather
than a reading, the system presents the candidates instead of choosing. The
judgments were retrieved from Find Case Law and used under a transactional
licence from The National Archives. Party names and case identifiers are not
reported: cases are referred to as A--F, and examples are given as relative
offsets rather than calendar dates. Neither the judgment texts nor any
judgment-derived text are released with the artifacts.

\section*{Artifact Availability}

The engine, graph representation, statutory rule packs and evaluation
harnesses are released as an installable tool under Apache-2.0 at
\url{https://github.com/mzhirko/timebar}. The research artifacts, including
the per-case results matrix, are at
\url{https://github.com/mzhirko/legal-temporal-dependency-graphs}. Neither
repository contains tribunal judgments.

\section*{Acknowledgments}

Crown copyright material reproduced by permission of The National Archives. The contents of the judgment can be used under the Open Justice -- Licence. The judgments used in this study only partially represent the activities of the courts and tribunals. Legislation is from legislation.gov.uk and contains public sector information licensed under the Open Government Licence v3.0.

We thank Coders Beyond Borders and their volunteer legal professional for reviewing the pilot output of the released tool and for a detailed verification report.

\bibliography{custom}

\appendix
\clearpage
% appendix_srw.tex -- matched to aacl-srw-draft-v7.tex
% Included from the main file via \appendix \clearpage \input{appendix_srw}

\section{Extended Related Work}
\label{app:related}

The work also connects to language models as reasoners and autoformalisers, to
legal rule representation, and to executable law. Chain-of-thought outputs
expose intermediate steps, and our direct baseline shows that an exposed
working does not guarantee a faithful final field. Autoformalisation systems
require independent compilation or execution, because generated syntax that
compiles can still compute the wrong value. The TDG and Catala tracks are
complementary. The graph carries document-level provenance and cross-document
dependencies, and Catala checks executable calendar rules.

\section{Data, Annotation, and Preprocessing}
\label{app:data}

\paragraph{Contracts.} Fifty seeds were sampled from the \texttt{en\_contracts}
subset. This returned 45 documents, of which 41 are unique under exact hashing
of the source text. The duplicates are one document appearing three times and
two appearing twice, and they are not evenly weighted: the triplicate carries 89
rule-based facts, so retaining duplicates inflates the rule-based fact total by
roughly a third. All extraction figures in the paper are computed over the 41
unique documents.

\paragraph{Verdict-level gold.} Full TDG annotation is expensive and is not
required for the deadline task. Verdict-level gold records the statutory anchor,
the claim presentation date, the available ACAS certificate dates, the tribunal
verdict, and the deadline or boundary date stated by the judge. Two of the seven
rows carry a judge-stated deadline and one carries a judge-stated boundary date.
Every field is linked to a verbatim quotation and byte-checked against the
source text by script. One case states an in-time conclusion without reciting
the conciliation dates that produced it. It is marked document-bound throughout
and is the single miss in the engine's upper bound.

\paragraph{Preprocessing.} Contract PDFs contain merged columns, broken words,
and headers and footers inside body paragraphs. A structural cleaner is applied
for the rule-based pipeline. The LLM receives raw text, because the cleaner was
aggressive enough to remove genuine clause content. Statutes arrive as clean
HTML. Judgments are converted from National Archives XML, and redaction is
applied as a separate audited step with its own report.

\section{Extraction and Prompt Details}
\label{app:extraction}

The rule-based pipeline runs temporal extraction, role classification, entity
linking and graph construction. Governing-verb lemmas classify \texttt{START},
\texttt{END} and \texttt{DURATION} roles, with special handling for legal
genitives such as \textit{Decision of 18 March 1992}, which have no governing
verb but denote a start event. Coreference links pronouns to entities. The graph
builder compares the calendar gap of each shared-entity \texttt{START}--
\texttt{END} pair against every stated duration in the document. A matching gap
produces an additive edge carrying \texttt{delta\_days}, and an unmatched pair
produces an ordering edge.

HeidelTime resolves incomplete expressions such as \textit{by 31 March} against
a document creation time. A contract corpus supplies none, and a fixed global
value places such dates outside the document's own period. In one 1984
instrument it produced a date in 2020. We therefore derive a per-document
creation time from the document's earliest explicit date, ignoring bare years,
because citation years are not document dates. This resolves 43 of the 45
documents. The two carrying no date at all produce no edges.

The LLM prompt defines JSON schemas for facts and relations, supplies trigger
phrases for additive dependencies, includes a worked mapping from \textit{within
30 days of the effective date} to an edge with \texttt{delta\_days: 30}, and
gives counterexamples that prevent monetary amounts and percentages from being
typed as durations. Post-processing removes self-loops and downgrades additive
claims that no trigger phrase supports. The model also emits \texttt{interval}
edges, which in this corpus connect nodes with no date and therefore fall
outside the computable set.

\subsection{Per-edge audit}
\label{app:edge-audit}

Both computable edge sets are small enough to audit exhaustively. Each edge was
read against the source sentences of both endpoints and judged \emph{correct}
(offset and anchor both stated in the text and correctly identified),
\emph{wrong anchor} (offset real, attached to the wrong event), or
\emph{spurious}. The scoring rule was fixed before the audit and applied
identically to both extractors.

\begin{table}[h]
\centering
\small
\begin{tabular}{llr}
\toprule
Doc & Edge & Verdict \\
\midrule
\multicolumn{3}{l}{\emph{Rule-based (9 computable)}}\\
A & \texttt{P3M} from 15 Mar 1986 & correct \\
A & \texttt{P3Y}, ``last three years'' & spurious \\
B & \texttt{P1Y}, secondment minimum & spurious \\
C & \texttt{P6M} vs stated one year & spurious \\
D & \texttt{P1M} from signature date & spurious \\
E & \texttt{P3Y} from wrong anchor & spurious \\
F & \texttt{P3Y} from funding schedule & spurious \\
F & \texttt{P30D}, unrelated clause & spurious \\
F & \texttt{P30D} from a bare year & spurious \\
\addlinespace
\multicolumn{3}{l}{\emph{LLM (10 computable)}}\\
G & notify within one month & correct \\
G & notify within one month & correct \\
G & review within five years & correct \\
H & 21 days from entry into force & wrong anchor \\
I & 10-year initial period & spurious \\
I & 5-year renewal from signature & spurious \\
C & withdrawal notice as offset & spurious \\
C & \texttt{P180D}, unrelated clause & spurious \\
J & \texttt{P120D}, ``at any time'' & spurious \\
K & anchor carries no date & spurious \\
\bottomrule
\end{tabular}
\caption{Every computable edge, audited. The three correct LLM edges share one
shape: an explicit trigger phrase naming the anchor concept in the same
sentence.}
\label{tab:edge-audit}
\end{table}

One LLM edge disagrees with itself in three places. Its expression reads
\textit{three months}, its endpoint carries a ninety-day duration, and its
offset field records thirty. It is the only arithmetic inconsistency in either
extractor.

\section{Controlled Benchmark Construction}
\label{app:controlled}

The 210 items are generated with a fixed seed in seven balanced categories of
30. Ground truth is computed from the calendar at generation time, so no system
can agree its way to a wrong answer and the set can be regenerated exactly.

Anchors are drawn from a stress list chosen to exercise the cases that produce
most errors: 31 January, 29 and 30 January, 29 February 2024, and month ends
including 31 August and 30 November. Deadline items pick an anchor, a period and
a margin from $\{-7,-4,-3,-1,1,3,4,7\}$ days, compute the deadline, and place
the action date relative to it. The period is expressed as digits, as words, or
as an idiom such as \textit{a fortnight}. Cascade items pick a root date and a
chain of two or three offsets, compute the original chain, replace the root and
recompute, so the system must propagate the correction. Cross-document cascade
items restate the final event in a second document with a vague offset, so the
same event must receive one consistent corrected date.

Answers are compared by calendar value and not by surface form, so a correctly
computed date is not penalised for formatting. Format compliance is tracked
separately.

\begin{table}[h]
\centering
\small
\begin{tabular}{lrrr}
\toprule
Approach & 1st link & 2nd link & 3rd link \\
\midrule
Structured & 100 & 100 & 100 \\
gemma4:e4b & 97 & 96 & 93 \\
llama3.1:8b & 11 & 8 & 3 \\
\bottomrule
\end{tabular}
\caption{Per-link accuracy on the cascade task (\%), the share of individual
linked dates correct at each step. The structured method is exact at every step
by construction.}
\label{tab:perlink}
\end{table}

\section{Catala Verification}
\label{app:catala}

An LLM translates extracted rules into Catala and the compiler independently
recomputes the derived dates. Agreement is evidence that the extracted rule is
arithmetically self-consistent. It is not evidence that the source was read
correctly, because two representations encoding the same misread date will
agree and both be wrong.

Unlike the extraction figures elsewhere in this paper, which are computed over
the 41 unique documents, this comparison is reported over the 45 sampled
documents that reach the comparison stage, because the Catala track is run
per sampled document. Of the 45 generated programs, 40 compile and execute; two
fail the repair loop and three fail at interpretation.

The two tracks rarely describe the same fields. Across the 45 documents, 116
fields appear only in the graph and 23 only in the Catala output, and only 19
documents yield a field both sides describe. Of those 19, 16 yield a field where
both sides give a value of the same kind. Over that shared subset the comparator
records 14 agreements, 6 value mismatches, and 5 type mismatches where one side
gives a date and the other a duration, so there is nothing to compare. In one of
the six value mismatches the cross-check caught the TDG side misreading a
duration introduced by scanning errors in the source document.

The small overlap is the finding. Coverage is the binding limit: the two
representations model different slices of each document, so most fields are
produced by one track only, and we report this as complementarity and not as an
agreement rate.

\section{Deadline Computation Mechanics}
\label{app:mechanics}

Figure~\ref{fig:mechanics} shows one computation end to end, with every
quantity the engine reports alongside the verdict.

\begin{figure*}[h]
\centering
\resizebox{0.95\textwidth}{!}{%
\begin{tikzpicture}[
  font=\small,
  >={Stealth[length=2.2mm]},
  ev/.style={circle, draw, fill=white, inner sep=1.3pt, line width=0.6pt},
  evlbl/.style={font=\scriptsize, align=center},
  note/.style={font=\scriptsize\itshape, text=black!60, align=center}
]

\def\xanchor{1.3}
\def\xdayA{4.8}
\def\xdayB{7.2}
\def\xprim{8.9}
\def\xpres{11.3}
\def\xeff{13.1}

\draw[->, line width=0.7pt] (0,0) -- (15.6,0);

% ---- primary period
\fill[blue!12, rounded corners=1pt] (\xanchor,0.18) rectangle (\xprim,0.56);
\node[note, text=blue!45!black] at ({(\xanchor+\xprim)/2},0.95)
  {primary period: anchor $+$ 3 months $-$ 1 day};

% ---- extension
\fill[green!16, rounded corners=1pt] (\xprim,0.18) rectangle (\xeff,0.56);
\node[note, text=green!35!black] at ({(\xprim+\xeff)/2},1.32)
  {extension equals the length\\of the pause, with a floor of\\one month after Day B};
\draw[line width=0.4pt, green!35!black]
  ({(\xprim+\xeff)/2},0.62) -- ({(\xprim+\xeff)/2},0.95);

% ---- conciliation pause
\fill[orange!22, rounded corners=1pt] (\xdayA,-0.56) rectangle (\xdayB,-0.18);
\node[note, text=orange!55!black] at ({(\xdayA+\xdayB)/2},-0.95)
  {clock suspended};

% ---- events
\node[ev] (a)  at (\xanchor,0) {};
\node[ev] (dA) at (\xdayA,0)   {};
\node[ev] (dB) at (\xdayB,0)   {};
\node[ev] (pd) at (\xprim,0)   {};
\node[ev] (ps) at (\xpres,0)   {};
\node[ev, fill=green!30] (ed) at (\xeff,0) {};

\node[evlbl, above=3.5mm of a]  {\textbf{anchor}\\effective date\\of termination};
\node[evlbl, below=3.5mm of dA] {\textbf{Day A}\\contact\\with ACAS};
\node[evlbl, below=3.5mm of dB] {\textbf{Day B}\\certificate\\issued};
\node[evlbl, above=3.5mm of pd] {\textbf{primary}\\\textbf{deadline}};
\node[evlbl, below=3.5mm of ps] {\textbf{claim}\\\textbf{presented}};
\node[evlbl, above=3.5mm of ed] {\textbf{effective}\\\textbf{deadline}};

% ---- verdict
\node[note, text=green!35!black, anchor=west] at (14.0,-0.75)
  {presented before the\\effective deadline:\\\textbf{in time}};
\draw[->, line width=0.5pt, green!35!black] (\xpres,-0.30) -- (\xeff,-0.30);

% ---- precondition
\draw[<->, line width=0.5pt, black!45] (\xanchor,-2.45) -- (\xprim,-2.45);
\node[note] at ({(\xanchor+\xprim)/2},-2.85)
  {precondition: Day A must fall inside the primary period,\\otherwise no extension applies};

\end{tikzpicture}%
}
\caption{Mechanics of one deadline computation. The primary period runs from
the anchor for the statutory period, less one day, under the \textit{beginning
with} convention. Early conciliation suspends the clock between Day A and Day
B, and the suspended interval is added to the primary deadline, with a floor of
one month after Day B. The engine also checks that Day A falls inside the
primary period, since contact made after the deadline has passed grants no
extension. Every quantity shown is reported alongside the verdict.}
\label{fig:mechanics}
\end{figure*}

\section{Matcher Recall and Ablation}
\label{app:matcher}

A recall audit of the 28 extracted TDGs shows that the facts the engine needs
are largely present and that the binding constraint lies downstream.

\begin{table}[h]
\centering
\small
\begin{tabular}{lrrrr}
\toprule
Condition & Anchor & Day A & Day B & Presented \\
\midrule
gemma / redacted & 4/7 & 3/4 & 2/3 & 7/7 \\
gemma / fulltext & 5/7 & 3/4 & 2/3 & 7/7 \\
mini / redacted & 5/7 & 4/4 & 3/3 & 7/7 \\
mini / fulltext & 6/7 & 4/4 & 3/3 & 7/7 \\
\bottomrule
\end{tabular}
\caption{Extraction recall of gold key dates per pipeline condition,
deduplicated per case.}
\label{tab:recall}
\end{table}

The engine, its counting conventions and the rule specifications were fixed
before the tribunal evaluation. The matcher uses five statute-grounded
mechanisms: anchor aliases derived from statutory vocabulary, a stoplist for
procedural events, claim-presentation cues with an earliest-date tie-break, a
conflict gate for disagreeing anchor candidates, and tiered matching so that
weak lexical similarity cannot outrank statutory terms. An ablation re-running
the same TDGs, with no re-extraction and no new model calls, shows that these
mechanisms raise correct matches from 6 of 10 to 12 of 14 across the affected
tables and reduce abstentions.

Across the 28 cells the engine answers 14 and abstains on 14. Twelve of the 14
answers are correct verdicts, and both errors are the same document-bound case.
Verdict agreement alone is not sufficient, and the main paper reports the
stricter figure: 9 of the 12 correct verdicts also reproduce the day count, and
3 are correct by margin. Two of those three are the same case under two
conditions, where the anchor is one day before the gold anchor and the deadline
follows it one day early, against a margin of 61 days. The third binds the anchor to the end of a harassment period rather than
the end of the continuing act, 206 days too early, and takes a grievance for
the claim, dated 239 days before the actual presentation date, and returns the
right verdict from two wrong inputs. All values are read from the archived
replay, in which four matcher variants over the same 28 extracted graphs return
identical answers on every cell.

The residual cases expose two kinds of limit. Extracted labels can import the
same rejected-contention error seen in direct reasoning. And identifying the end
of a continuing act can itself be a legal judgment: the appellate court
described the tribunal's scoping as clearly open to it, which makes the models'
contrary view arguable. The conflict gate therefore surfaces the candidate
computations instead of selecting one silently.

\section{Counterfactual Harness}
\label{app:counterfactual}

\paragraph{Why the naive design fails.} Rewriting only the anchor date's own
surface forms leaves its satellites in place. A judgment that elsewhere says
\textit{his summary dismissal in} followed by the month and year, or dates a dismissal letter the day
before and service two days after, becomes self-contradictory about when the
anchor occurred. On such text a conflict-gated matcher abstains correctly and a
direct model anchors silently on the stale mention. Both behaviours measure the
corruption of the test set and not the system. An earlier run on such texts is
retained as a validity study. Its abstentions concentrated on the items whose
documents contained contradictory anchor evidence, while the direct models
answered the same self-contradictory documents at full coverage without
signalling a problem.

\paragraph{Construction.} The final harness translates the past. Every date
expression falling before the first oracle-frozen event, which is the first
conciliation date or the presentation date where no conciliation applies, is
shifted by $k$, so the anchor and its narrative move as one block while the
conciliation and litigation timeline stays fixed. Full dates, bare day-months
and bare month-years are all rewritten. A residual-reference detector scans
every emitted text for surviving partial references to the original date, and
generation enforces a zero-residual gate. Bare year tokens are not rewritten,
because citation years such as \textit{[2018] EAT} make that unsafe, and they
are counted and flagged per item instead.

For each gold case the anchor is swept across $k \in [b-30, b+30]$, where $b$ is
the engine-computed boundary offset for that case, so every sweep straddles its
own statutory boundary. This gives $7 \times 61 = 427$ items, balanced 31/30
timely to late within every case. A self-check verifies that each sweep is
monotone with exactly one verdict flip, at the boundary the statute prescribes.
The engine generates the labels and is not scored against them, so the
experiment measures text-reading systems against statute-computed truth.

Cleaning the texts raised the \emph{direct} models as well as the pipeline, from
59.0\% to 64.6\% and from 58.8\% to 60.2\%, which indicates that the
regeneration was a validity fix and not an accommodation of our own method.

\begin{table}[h]
\centering
\small
\begin{tabular}{lrrr}
\toprule
Condition & Anchor-correct & Flips & One flip \\
\midrule
gemma direct & --- & 68 & 1/7 \\
llama direct & --- & 117 & 0/7 \\
gemma pipeline & 0.983 & 24 & 3/7 \\
llama pipeline & 0.939 & 18 & 0/7 \\
\bottomrule
\end{tabular}
\caption{Boundary behaviour on the 427 items. Anchor-correct is accuracy on the
subset where the extracted anchor equals the true shifted anchor ($n=234$ and
$n=132$). A system tracking the statute flips exactly once per sweep. On rows
where the extracted anchor is wrong the pipeline sits at chance (0.538 and
0.452), which is the behaviour a calibrated system should show when its input is
wrong.}
\label{tab:cf-detail}
\end{table}

\section{Consistency Battery}
\label{app:coherence}

Six questions are asked per case: the presentation date, the effective deadline,
the last date on which the claim could still have been in time, the verdict, the
number of days late, and whether presentation complied. Four constraints hold by
arithmetic or definition. K1: the last permissible date equals the deadline. K2:
the day count equals presentation minus deadline. K3: the verdict agrees with
the sign of the day count. K5: the compliance flag agrees with the verdict. A
fifth relation, that the verdict agrees with presentation compared against the
last permissible date, is entailed by K1 to K3 and is computed only as a
redundancy check. Counting it alongside its own premises would inflate the
independent tests per case from four to five.

K5 is a control and not a fourth arithmetic test. It asks the same question as
K3 in different words and with a different answer type, so a violation confined
to K5 indicates sensitivity to phrasing and not an arithmetic failure.

\paragraph{Three populations.} Coverage counts the cases where all six answers
parsed, and the incoherence rate is computed over those. Verdict accuracy is
computed over the rows whose verdict field parsed, which is a larger set. The
per-case coherence flag is evaluated over whatever constraints are evaluable, so
a partial battery can still be found contradictory. These three populations are
not nested and must not be mixed. The published rate uses the first.

\begin{table}[h]
\centering
\small
\begin{tabular}{lrrrr}
\toprule
Condition & K1 & K2 & K3 & K5 \\
\midrule
llama, separate & 5/6 & 3/4 & 3/7 & 1/7 \\
llama, single & 4/6 & 6/6 & 3/6 & 0/6 \\
gemma, separate & 3/6 & 4/7 & 1/6 & 1/6 \\
gemma, single & 0/7 & 0/7 & 0/7 & 0/7 \\
gpt-5.4, separate & 2/7 & 4/7 & 0/7 & 0/7 \\
gpt-5.4, single & 0/7 & 0/7 & 0/7 & 0/7 \\
\bottomrule
\end{tabular}
\caption{Constraint violations over evaluable cases. K1 dates, K2 arithmetic,
K3 verdict, K5 paraphrase control. The frontier model's failures are entirely
K1 and K2, so its reporting layer never contradicts its own day count.}
\label{tab:coherence-constraints}
\end{table}

In the single-prompt condition llama violates K2 in 6 of 6 evaluable cases and
K5 in none. With both operands present in its own emitted output it performed
the subtraction wrongly every time, while never contradicting its own
conclusion.

\paragraph{Coverage is not missing at random.} No condition contains an
abstention, and every coverage loss is a parse failure. All of them fall on
three cases: the two whose anchors are hardest elsewhere in our results, and one
whose anchor is a rejected contention. Accuracy over answered items is therefore
computed on an easier subset.

\section{Failure Taxonomy}
\label{app:taxonomy}

The smaller models fail the minus-one-day convention, the conciliation
preconditions and anchor selection. Some fabricate missing certificate dates or
assume the judgment's conclusion. One produced a correct primary deadline
through two errors that cancelled, which is the clearest single argument for
auditing the working and not the value. The frontier model solves most of these
arithmetic issues and can still emit a verdict or an effective deadline
inconsistent with its own working.

A separate leakage class occurs when a model defers to the direction of the
appeal despite its own arithmetic pointing the other way. Two models computed
\emph{out of time} and then deferred to what the judgment appeared to indicate.
We treat the verdict metric as contaminated on that case for models that defer.
Deadline-error metrics are unaffected, because no deadline value survived
redaction.

All four models scope one continuing act the same way, differing from the
tribunal by 206 days. This is a divergence of legal judgment and not an
arithmetic error, it does not change with model scale, and it is the shared
boundary of all automation here including our own, whose gold anchor is
judge-found.

The complete per-case matrix is in the research repository (see Artifact
Availability), because aggregate verdict
accuracy cannot distinguish these mechanisms.

\section{Additional Tasks and Negative Results}
\label{app:negative}

\paragraph{Cross-document contradiction detection.} A detector was built and
tested on document pairs with planted conflicts. On the development set it
detects five of five with no false positives, after three changes to the linker
that introduce no tuned constants. The first treats a gap between two stated
periods as the disputed quantity and not as identity evidence. The second
requires a dispute to compare two dates or two periods, because a date and a
duration are two encodings of one fact and not a disagreement. The third treats
sentences identical token for token except at value positions as the same clause
by construction, so weak entity naming cannot veto a match.

On a mechanically planted held-out set, generated by a committed seeded script
and scored blind, the detector finds one of four in each of three extraction
repeats. Every miss traces above the linker, to extraction: one event is never
extracted at all, and another is named in Dutch in one run and in English in the
next.

We report this as a bounded result. Cross-document contradiction detection is
limited by entity-naming consistency, and the boundary between a contradiction
and two provisions lawfully applying in parallel is itself ill-defined for
independent documents. We encountered this in real data, where two review
deadlines differ and each is true of its own protocol.

\paragraph{Pairwise ordering.} On a legal event-ordering benchmark
\cite{barale-etal-2025-lextime} the
structured route loses outright. Coverage falls to 18\% and 28\% because most event pairs
are never connected by an extracted relation, and direct answering wins.
Ordering has no arithmetic to verify, so the method pays its extraction cost and
receives nothing back. This marks the boundary of the claim from the outside, as
the deadline results mark it from the inside.

The same benchmark reproduces the consistency result on public data. It contains
190 event pairs asked in both directions, of which exactly one direction can be
true. Giving the same answer to both is a self-contradiction whatever the
correct ordering, and the direct models do so on 32\% and 52\% of pairs
($61/190$ and $99/189$), against 0\% for the structured route on the 41 and 54
pairs it answers, where a single derivation decides both. It also shows that
item-level accuracy overstates ordering competence, because a self-contradictory
pair scores 50\% instead of 0\%, worth 12 and 22 points respectively.

\end{document}